\documentclass[runningheads]{llncs}
\usepackage[T1]{fontenc}
\usepackage{graphicx}
\usepackage{multirow}
\usepackage{tikz}
\usepackage{comment}
\usepackage{amsmath,amssymb} 
\usepackage{color}
\usepackage{float}
\usepackage{marvosym}
\usepackage{paralist}
\usepackage{enumitem}
\usepackage{colortbl}
\usepackage{booktabs}
\usepackage{subcaption}
\usepackage{hyperref}
\hypersetup{
    colorlinks=true,
    linkcolor=blue,
    filecolor=magenta,      
    urlcolor=red,
}
\newcommand{\etal}{\textit{et al.}}

\begin{document}
\title{Customized Segment Anything Model for Medical Image Segmentation}
%
%
\author{Kaidong Zhang \and
Dong Liu}
\authorrunning{Zhang and Liu}
%
\institute{University of Science and Technology of China\\
\email{richu@mail.ustc.edu.cn, dongeliu@ustc.edu.cn}\\}
\maketitle              
\begin{abstract}
We propose SAMed, a general solution for medical image segmentation. Different from the previous methods, SAMed is built upon the large-scale image segmentation model, Segment Anything Model (SAM), to explore the new research paradigm of customizing large-scale models for medical image segmentation. SAMed applies the low-rank-based (LoRA) finetuning strategy to the SAM image encoder and finetunes it together with the prompt encoder and the mask decoder on labeled medical image segmentation datasets. We also observe the warmup finetuning strategy and the AdamW optimizer lead SAMed to successful convergence and lower loss. Different from SAM, SAMed could perform semantic segmentation on medical images. Our trained SAMed model achieves 81.88 DSC and 20.64 HD on the Synapse multi-organ segmentation dataset, which is on par with the state-of-the-art methods. We conduct extensive experiments to validate the effectiveness of our design. Since SAMed only updates a small fraction of the SAM parameters, its deployment cost and storage cost are quite marginal in practical usage. The code of SAMed is available at \url{https://github.com/hitachinsk/SAMed}.

\keywords{Medical image segmentation  \and SAM \and Finetune.}
\end{abstract}

\begin{figure*}[t]
\begin{center}
\includegraphics[width=\linewidth]{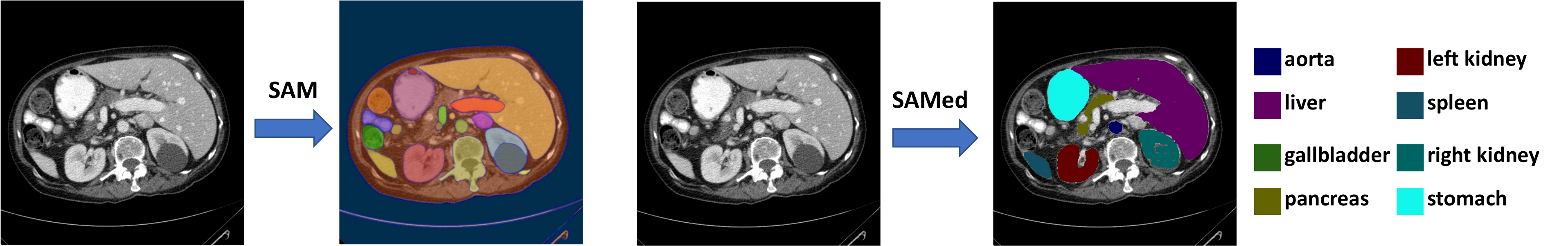}
\end{center}
   \caption{SAMed inherits the remarkable image segmentation performance from SAM and further refines the segmentation boundaries with the anatomical prior knowledge from the professional medical data during customization process. Moreover, SAMed can fully understand the semantic class of each segmentation regions by classifying these regions to different meaningful tissues for automatic medical image semantic segmentation.}
\label{fig:teaser}
\end{figure*}

\section{Introduction}
Medical image segmentation aims at indicating the anatomical or pathological structure of the corresponding tissues as required to facilitate computer-aided diagnosis and intelligent clinical surgery~\cite{lei2020medical}. Owing to the rapid expansion of computation power and medical data resources, deep learning based medical image segmentation have achieved massive progress in accuracy and speed against traditional counterparts.

Compared with CNN-based methods~\cite{ronneberger2015u,jin2020ra,schlemper2019attention}, the integration of transformer blocks~\cite{chen2021transunet,swinunet,10.1007/978-3-031-16919-9_9,azad2023daeformer} makes it possible to attend and aggregate similar features in global manner, which leads to unprecedented performance in medical image segmentation. However, designing such networks towards a specific dataset requires massive amount of network engineering. What's more, the overhead in deploying and storing these models for a specific usage is non-negligible due to their considerable model size, which presents a huge challenge towards the practical usage.

Nowadays, AI research community is experiencing a dramatic revolution. A proliferation of large-scale models, such as DALL-E~\cite{ramesh2021zero}, GPT-4~\cite{openai2023gpt4}, SAM~\cite{kirillov2023segment} and SegGPT~\cite{SegGPT} make it possible for researchers to solve multiple kinds of problems in a unified framework. Deploying such large-scale models is more promising on the purpose of industrial usage because of their remarkable generalization capability. In the field of medical image segmentation, if large-scale CV models, such as SAM or SegGPT can achieve highly competitive performance, it will be not necessary to deploy a single medical image segmentation model and the solution to medical image segmentation can be integrated into the large-scale CV models directly. Moreover, the burden of network engineering for medical image segmentation will be not necessary and the deployment and storage overhead of an individual model for medical image segmentation can also be greatly saved.

Unfortunately, due to the lack of medical image data and their corresponding semantic labels, the large-scale CV models cannot be utilized to solve medical image segmentation directly. First, the large-scale CV models make the decisions of boundaries between different segmentation regions based on the variance of intensity~\cite{ji2023sam,MedSAM}, which is reasonable in natural images but not medical images because the analysis of anatomical or pathological structure plays a critical role in medical image segmentation. Second, the large-scale CV models are incapable of associating the segmentation regions to meaningful semantic classes. In other words, they cannot perform semantic segmentation for medical images, which hinders their usage in computer-aided diagnosis.

Based on the above analysis, we explore the methods to customize one of the representative large-scale CV models - SAM (Segment Anything Model) for medical image segmentation and conclude some useful strategies to improve its performance. Our method, SAMed (\textbf{S}egment \textbf{A}nything Model for \textbf{Med}ical), is built upon SAM and only add and update a small fraction of parameters during the customization process, which means the deployment and storage of SAMed is marginal on the segment anything system. What's more, SAMed can perform semantic segmentation for medical images and achieve remarkable performance. We illustrate the difference of the segmentation results in Fig.~\ref{fig:teaser}. Technically, We freeze the image encoder and adopt low-rank-based finetuning strategy (LoRA)~\cite{hulora} to approximate the low rank update of the parameters in image encoder, and finetune the lightweight prompt encoder and the mask decoder of SAM. In ``vit\_b" mode, the updated model size (18.81M) only occupies 5.25\% of the original model size (358M). If we apply LoRA to both of the image encoder and mask decoder, the model size will get further reduced to 6.32M but the performance slightly drops. In terms of training strategies, we observe warmup~\cite{he2016deep,xiong2020layer} and AdamW~\cite{loshchilovdecoupled} optimizer could greatly stabilize the finetuning process, which leads to higher segmentation precision. After finetuning only 160 epochs on Synapse multi-organ segmentation dataset (Synapse), SAMed achieves 81.88 DSC and 20.64 HD, which is on par with the current state-of-the-art (SOTA) baselines. SAMed can be regarded as a plugin of SAM and is fully compatible with SAM. During inference, we just switch the updated layers of SAMed to empower SAM with the ability to process medical images.

We summarize the contributions of this paper as:
\begin{itemize}
    \item We firstly extend SAM to explore its capability on medical image segmentation with semantic labels.
    \item We present the adaptation of image encoder and a series of finetuning strategies in the consideration of performance and deployment and storage overhead.
    \item Our method, SAMed, achieves highly competitive results against previous well-designed medical image segmentation methods in both DSC and HD.
\end{itemize}

\section{Related Works}
\subsection{Medical image segmentation models} Early medical image segmentation methods empirically adopt explicit contour features~\cite{1194625} or Markov random field~\cite{held1997markov} to obtain decent performance. With the rapid progress of deep learning, U-Net~\cite{ronneberger2015u}, as the seminal work, initiates a new era for medical image segmentation. Following U-Net, researchers design multiple variants to increase the performance of medical image segmentation, including Res-UNet~\cite{xiao2018weighted}, Dense-UNet~\cite{li2018h}, U-Net++~\cite{zhou2019unetplusplus}, 3D-Unet~\cite{cciccek20163d}, etc. Recently, the application of Transformer~\cite{vaswani2017attention} in nature image also sparks the research in medical image segmentation. The mainstream network design strategy is to integrate the transformer blocks to the U-Net framework, including TransUnet~\cite{chen2021transunet}, SwinUnet~\cite{swinunet}, MiSSFormer~\cite{9994763}, Hi-Former~\cite{heidari2023hiformer}, DAE-Former~\cite{azad2023daeformer}, etc. Compared with these methods, SAMed does not require elaborate network engineering, and can still achieve highly competitive performance. A concurrent work, MedSAM~\cite{MedSAM}, also adapts SAM for medical image segmentation. Different from MedSAM, SAMed can perform semantic segmentation for medical images, and the adaptation of image encoder also helps SAMed to extract the specific features for more accurate medical image segmentation.

\subsection{Large-scale models} The remarkable extension ability of Transformer makes it possible to construct large-scale models with billions of parameters. Large-scale models firstly make breakthrough in natural language processing, such as BeRT~\cite{kenton2019bert}, InstructGPT~\cite{ouyang2022training}, LLaMA~\cite{touvron2023llama}, GPT-4~\cite{openai2023gpt4}, etc. They all achieve awesome performance against previous methods. Nowadays, large-scale model in computer vision. including SAM~\cite{kirillov2023segment}, SegGPT~\cite{SegGPT}, STU-Net~\cite{huang2023stu} and their extended applications receive a huge amount of attention. Although these large-scale models equip with astonishing zero-shot generalization ability, they fail to understand and provide precise anatomical structure of the target tissues due to the lack of professional data. SAMed designs specific strategies to customize SAM for medical image segmentation with semantic labels while balancing segmentation accuracy, deployment and storage overhead.

\subsection{Finetuning strategies} Pretrained large-scale models could learn and extract excellent features due to the high quality learning from large-scale data. Therefore, how to utilize the ability of large-scale models and instruct the downstream tasks with new knowledge becomes a valuable problem. Current solutions tend to improve the finetuning strategies to inject the new knowledge to the pretrained large-scale models. Compared with finetuning all the parameters in large-scale models, visual prompt tuning~\cite{jia2022vpt} proposes to finetune the prompt and the head to achieve excellent performance while reducing the training cost. LoRA~\cite{hulora} argues the update of parameters in transformer blocks is gradual and designs the low rank approximation to finetune the large-scale language model. SAMed also adopts LoRA to customize SAM for medical image segmentation and explores several variants to design the LoRA in SAM to strike the balance between performance and efficiency.

\begin{figure*}[t]
\begin{center}
\includegraphics[width=\linewidth]{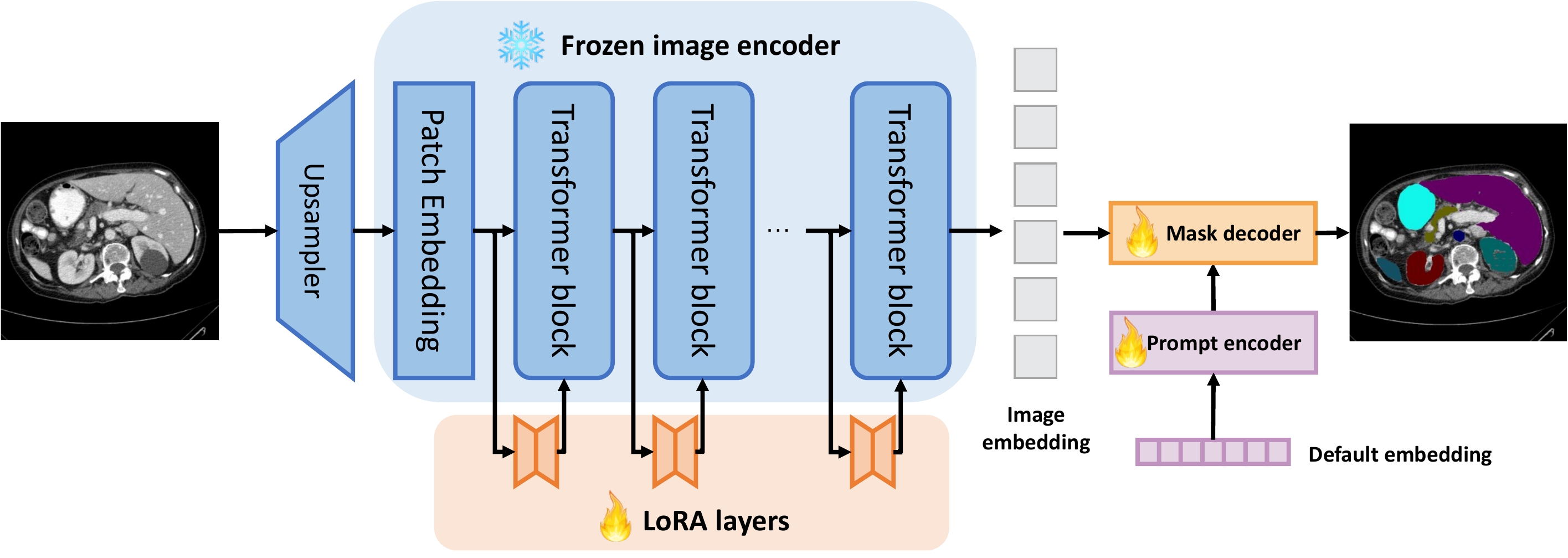}
\end{center}
   \caption{The pipeline of SAMed. The framework of SAMed is consistent with SAM. We freeze the image encoder, and insert additional trainable LoRA layers to SAM for medical image feature extraction. Moreover, we finetune prompt encoder with default embeddings and mask decoder to achieve precise semantic segmentation on medical images.}
\label{fig:pipeline}
\end{figure*}

\section{Method}

\subsection{Overview} Given a medical image $x\in \mathbb{R}^{H\times W\times C}$ whose spatial resolution is $H\times W$ and channel number is $C$, respectively, our goal is to predict its corresponding segmentation map $\hat{S}$ with resolution $H\times W$ where each pixel belongs to an element in a predefined class list $Y=\{y_0, y_1, ..., y_k\}$ as close to the ground truth $S$ as possible. We regard $y_0$ as the background class and $y_i, i\in \{1, ..., k\}$ as the classes of different organs. As illustrated in Fig.~\ref{fig:pipeline}, the overall architecture of SAMed inherits from SAM. We freeze all the parameters in the image encoder and design an trainable bypass to each of the transformer block. As indicated in LoRA, these bypasses condense the transformer features to the low rank space first and reproject the squeezed features to align with the channels of the output features in the freezed transformer blocks. As for the prompt encoder, SAMed does not need any prompt during inference to perform automatic segmentation, which greatly benefits automatic medical diagnosis. We note if we strip away all the prompts in SAM, SAM will update a default a default embedding, therefore SAMed also finetunes this embedding during training. The mask decoder of SAM can be roughly categorized into a lightweight transformer decoder and a segmentation head. Finetuning the transformer decoder with LoRA is optional in SAMed. If we freeze the transformer decoder and finetune it with LoRA layers but not finetune all its parameters, we can further shrink the model size of the updated parameters for easier deployment but the performance will drop slightly. The original segmentation head of SAM outputs multiple segmentation masks to solve the ambiguity in segmentation prompt, which is crutial in prompt engineering. To align with the original design of SAM, SAMed also predicts multiple segmentation masks, but each mask represents one class in $Y$, therefore SAMed predicts $k$ segmentation masks. Following SAM, the predicted segmentation logit size $h\times w$ is smaller than the input size, therefore we utilize bilinear upsample to align it with the original input size.

\begin{figure*}[t]
\begin{center}
\includegraphics[width=0.8\linewidth]{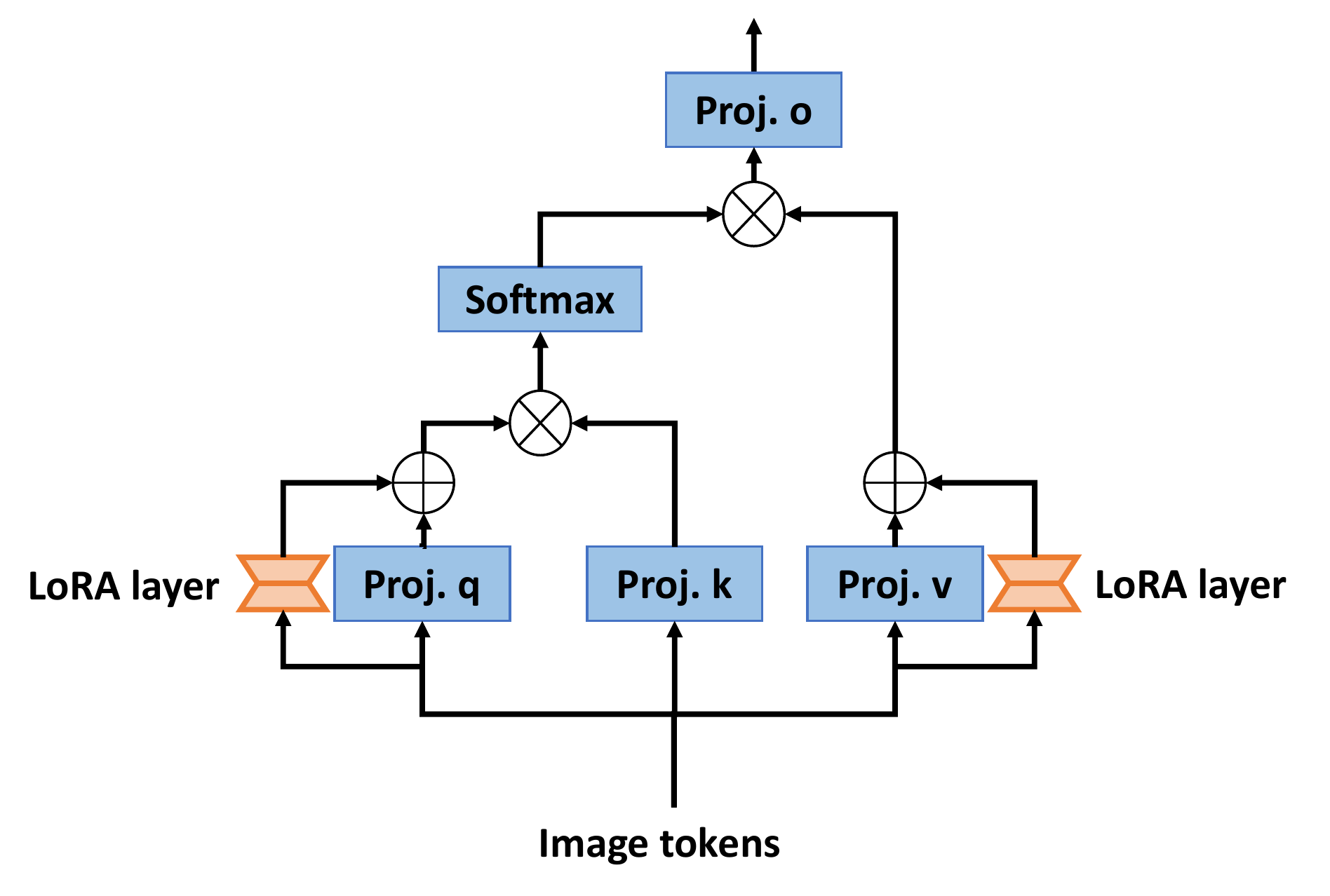}
\end{center}
   \caption{The LoRA design adopted in SAMed. We apply LoRA layer to the q and v projection layers of each of the transformer block in image encoder. ``Proj.q", ``Proj.k", ``Proj.v" and ``Proj.o" represent the projection layer of q, k, v and o, respectively.}
\label{fig:lora}
\end{figure*}

\subsection{LoRA in image encoder} Compared with finetuning all the parameters in SAM, LoRA allows SAM to update a small fraction of parameters during training on medical image, which not only saves computation overhead but also reduces the difficulty in deployment and storage of the finetuned models while guaranteeing the segmentation performance simultaneously. The LoRA strategy in SAMed is illustrated in Fig.~\ref{fig:lora}. Given the encoded token sequence $F\in \mathbb{R}^{B\times N\times C_{in}}$ and the output token sequence $\hat{F}\in \mathbb{R}^{B\times N\times C_{out}}$ operated by a projection layer $W\in \mathbb{R}^{C_{out}\times C_{in}}$, LoRA assumes the update of $W$ should be gradual and stable, therefore it proposes to apply low rank approximation to delineate this gradual update. Following this strategy, SAMed first freezes the transformer layers to keep $W$ fixed, and then adds a bypass to accomplish the low rank approximation. This bypass contains two linear layers $A\in \mathbb{R}^{r\times C_{in}}$ and $B\in \mathbb{R}^{C_{out} \times r}$, where $r \ll \min\{C_{in}, C_{out}\}$. Therefore, the processing of the updated layer $\hat{W}$ can be described as
\begin{equation}
\begin{aligned}
    \hat{F} &= \hat{W}F, \\
    \hat{W} &= W+\Delta W=W+BA.
\end{aligned}
\end{equation}

Since the multi-head self attention mechanism determines the regions to focus on with cosine similarity, it's sensible to apply LoRA to the projection layers of query, key or value to influence the attention scores. We observe SAMed can achieve better performance when we apply LoRA to the query and value projection layers, therefore the processing strategy of multi-head self attention will become
\begin{equation}
\begin{aligned}
    {\rm Att}(Q,K,V) &= {\rm Softmax}(\frac{QK^T}{\sqrt{C_{out}}}+B)V, \\
    Q &= \hat{W_q}F=W_qF+B_qA_qF, \\
    K &= W_k F, \\
    V &= \hat{W_v}F=W_vF+B_vA_vF.
\end{aligned}
\end{equation}

Where $W_q$, $W_k$ and $W_v$ are freezed projection layers from SAM, and $A_q$, $B_q$, $A_v$ and $B_v$ are trainable LoRA parameters.

\begin{figure*}[t]
\begin{center}
\includegraphics[width=\linewidth]{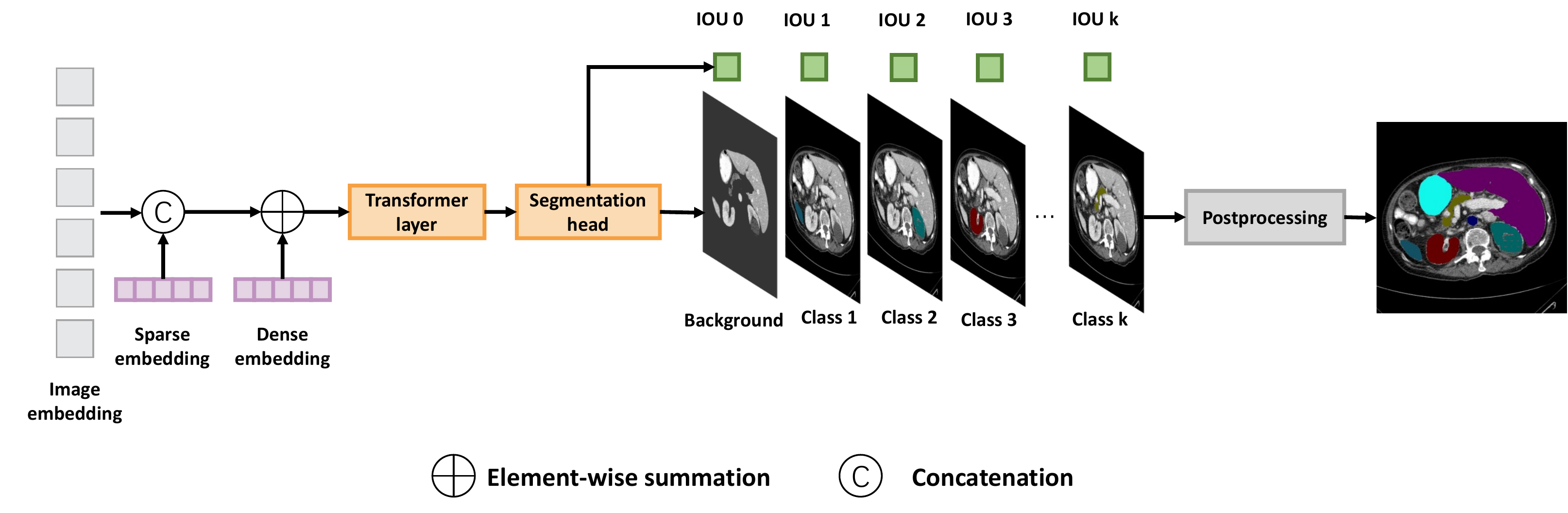}
\end{center}
   \caption{The detailed framework of mask decoder. SAMed integrates the sparse and dense embedding into the encoded image embedding. After processing by transformer layer, the segmentation maps together with their IOUs of each classes are generated individually. We adopt postprocessing to aggregate these segmentation maps to the final segmentation result.}
\label{fig:mask_decoder}
\end{figure*}

\subsection{Prompt encoder and mask decoder} To pursue fast and automatic medical diagnosis, SAMed does not need any prompt during inference. Since the prompt encoder in SAM utilizes a default embedding when there is not prompt provided, SAMed retains this default embedding and makes it trainable during finetuning process.

The mask decoder in SAM consists of a lightweight transformer layer and a segmentation head. It's optional to apply LoRA to this lightweight transformer layer and finetune the segmentation head or finetune all the parameters in the mask decoder directly. These two strategies are all acceptable in terms of training and deployment overhead, and the latter could lead to a smaller model size for easier deployment but lower performance. SAMed modifies the segmentation head of SAM slightly to customize the output for each semantic classes in $Y$. The customization is illustrated in Fig.~\ref{fig:mask_decoder}. Different from the ambiguity prediction of SAM, SAMed predicts each semantic class of $Y$ in a deterministic manner. Assume there are $k$ classes for segmentation, including 1 background class and $k-1$ classes corresponding each meaningful medical tissues. The mask decoder of SAMed predicts $k$ semantic masks $\hat{S_l}\in \mathbb{R}^{h\times w \times k}$ simultaneously, corresponding to each semantic label. Finally, the predicted segmentation map is generated as
\begin{equation}
    \hat{S} = {\rm argmax}({\rm Softmax}(\hat{S_l}, d=-1), d=-1).
\end{equation}

Where $d=-1$ indicates the $\rm Softmax$ and $\rm argmax$ operation are performed across the last dimension (the channel dimension).

\subsection{Training strategies}
\subsubsection{Loss function.} SAMed adopts cross entropy and dice losses to supervise the finetuning process. The loss function can be described as
\begin{equation}
    L = \lambda_1 {\rm CE}(\hat{S_l}, D(S)) + \lambda_2 {\rm Dice}(\hat{S_l}, D(S)).
\end{equation}

Where ${\rm CE}$ and ${\rm Dice}$ represents cross entropy loss and Dice loss, respectively. $D$ denotes as the downsample operation to make the resolution of the ground truth the same as that of the output from SAMed because its spatial resolution is much lower. $\lambda_1$ and $\lambda_2$ represent the loss weights to balance the influence between these two loss terms.

\subsubsection{Warmup.} Warmup is first proposed by ResNet~\cite{he2016deep} and its efficacy in transformer architecture is demonstrated by Xiong \etal~\cite{xiong2020layer}. SAMed adopts warmup to stabilize the training process and familiar with the medical data in the early training period. After the warmup period, following TransUnet~\cite{chen2021transunet} and SwinUnet~\cite{swinunet}, SAMed utilizes exponential learning rate decay to make the training process converge gradually. The learning rate adjustment strategy can be described as
\begin{equation}
\label{eq6}
lr=\left\{
\begin{aligned}
&T\frac{I_{lr}}{WP} , & T<=WP, \\
&I_{lr}(1-\frac{T-WP}{MI}) , & T>WP.
\end{aligned}
\right.
\end{equation}

Where $I_{lr}$ indicates the initial learning rate. $T$, $WP$ and $MI$ indicates the training iterations, warmup period and the maximal iterations, respectively.

\subsubsection{AdamW optimizer.} During experiments, we observe the AdamW~\cite{loshchilovdecoupled} optimizer leads to performance improvement compared with SGD in SAMed. We attribute this phenomenon to the consistency of the optimizing recipe between SAM and its improved version on medical image: SAMed.

\section{Experiments}

\subsection{Dataset and evaluation}
We adopt Synapse multi-organ segmentation dataset~\footnote{\url{https://www.synapse.org/\#!Synapse:syn3193805/wiki/217789}} for evaluation. 30 abdominal CT scans in the MICCAI 2015 Multi-Atlas Abdomen Labeling Challenge is divided into 18 training cases and 12 test cases. There are 3779 axial contrast-enhanced abdominal CT images in total and the training set contains 2212 axial slices. All the CT volumes contain 85$\sim$198 slices and each slices includes 512$\times$512 pixels with a spatial resolution of ([0.54$\sim$0.54]$\times$[0.98$\sim$0.98]$\times$[2.5$\sim$5.0]mm$^3$). The division between training and test cases is kept the same as TransUnet~\cite{chen2021transunet}. Following Fu \etal~\cite{fu2020domain}, we adopt the average DSC and the average Hausdorff distance (HD) on eight abdominal organs (aorta, gallbladder, spleen, left kidney, right kidney, liver, pancreas, stomach) as the metric to evaluate the performance of SAMed and other medical image segmentation models.

\subsection{Implementation details}
We adopt the same strategies of data augmentation as TransUnet~\cite{chen2021transunet} and conduct all the experiments based on the ``vit\_b" version of SAM. As for a 224$\times$224 CT image, we first upsample it to 512$\times$512 and then input this upsampled image to SAMed in order to maintain the decent image resolution of the predicted segmentation logits. The output resolution of segmentation logit for each class is 128$\times$128, which is smaller than that of UNet-based medical image segmentation models. There are 9 predicted segmentation logits, including one background class and 8 organ classes.

We adopt LoRA to finetune the freezed q and v projection layers of the transformer blocks. The rank of LoRA is set to 4 for efficiency and performance optimization. The loss weight of cross entropy is set to 0.2 and the that of Dice loss is set to 0.8. As for Warmup, we set the initial learning rate $I_{lr}$ to 0.005, the warmup period $WP$ to 250 and the maximal iterations to 18600 (200 epochs). $\beta_1$, $\beta_2$ and weight decay of AdamW optimizer are set to 0.9, 0.999 and 0.1. We adopt early stop at 14880 iterations (160 epochs).

\subsection{Comparison with SOTAs}

\begin{table*}[t!]
\centering
\caption{Quantitative comparison between SOTA methods and SAMed on the Synapse multi-organ CT dataset.}
\footnotesize
\resizebox{\textwidth}{!}{
\begin{tabular}{c|cc|cccccccc}
\toprule
Methods & DSC$\uparrow$ &HD$\downarrow$ & Aorta& Gallbladder& Kidney(L)& Kidney(R)& Liver& Pancreas& Spleen& Stomach\\
\midrule
U-Net~\cite{ronneberger2015u} & 76.85 & 39.70 & 89.07 & 69.72 & 77.77 & 68.60 & 93.43 & 53.98 & 86.67 & 75.58 \\
Att-UNet~\cite{oktayattention} & 77.77 & 36.02 & \textbf{89.55} & 68.88 & 77.98 & 71.11 & 93.57 & 58.04 & 87.30 & 75.75 \\
TransUnet~\cite{chen2021transunet} & 77.48 & 31.69 & 87.23 & 63.13 & 81.87 & 77.02 & 94.08 & 55.86 & 85.08 & 75.62 \\
SwinUnet~\cite{swinunet} & 79.13 & 21.55 & 85.47 & 66.53 & 83.28 & 79.61 & 94.29 & 56.58 & 90.66 & 76.60 \\
MissFormer~\cite{9994763} & 81.96 & 18.20 & 86.99 & 68.65 & 85.21 & \textbf{82.00} & 94.41 & 65.67 & 91.92 & 80.81 \\
TransDeepLab~\cite{10.1007/978-3-031-16919-9_9} & 80.16 & 21.25 & 86.04 & 69.16 & 84.08 & 79.88 & 93.53 & 61.19 & 89.00 & 78.40 \\
HiFormer~\cite{heidari2023hiformer} & 80.39 & \textbf{14.70} & 86.21 & 65.69 & 85.23 & 79.77 & 94.61 & 59.52 & 90.99 & 81.08 \\
DAE-Former~\cite{azad2023daeformer} & \textbf{82.43} & 17.46 & 88.96 & \textbf{72.30} & \textbf{86.08} & 80.88 & \textbf{94.98} & 65.12 & \textbf{91.94} & 79.19 \\
\midrule
SAMed & 81.88 & 20.64 & 87.77 & 69.11 & 80.45 & 79.95 & 94.80 & \textbf{72.17} & 88.72 & \textbf{82.06}\\
\bottomrule
\end{tabular}
}
\label{synapse}
\end{table*}

\subsubsection{Quantitative comparison.} We compare SAMed with current SOTA methods on Synapse dataset, including U-Net~\cite{ronneberger2015u}, Att-UNet~\cite{oktayattention}, TransUnet~\cite{chen2021transunet}, SwinUnet~\cite{swinunet}, MissFormer~\cite{9994763}, TransDeepLab~\cite{10.1007/978-3-031-16919-9_9}, HiFormer~\cite{heidari2023hiformer} and DAE-Former~\cite{azad2023daeformer}. The results are shown in Tab.~\ref{synapse}. We acknowledge the performance of SAMed is highly competitive but not the best among these SOTA methods, which demonstrates the SAM-based models are capable of achieving quite promising performance on medical image semantic segmentation with suitable customization strategies. Especially, SAMed achieves SOTA performance on the segmentation of pancreas and stomach. Different from the previous elaborate medical image segmentation methods that treat medical image segmentation as a total individual problem and require entire model deployment and storage, SAMed bridges the gap between nature image segmentation and medical image segmentation under a unified framework with the customization on only a small fraction of the parameters in SAM, which brings only marginal increment on the overhead of the original segment anything system.

\begin{figure*}[t]
\begin{center}
\includegraphics[width=\linewidth]{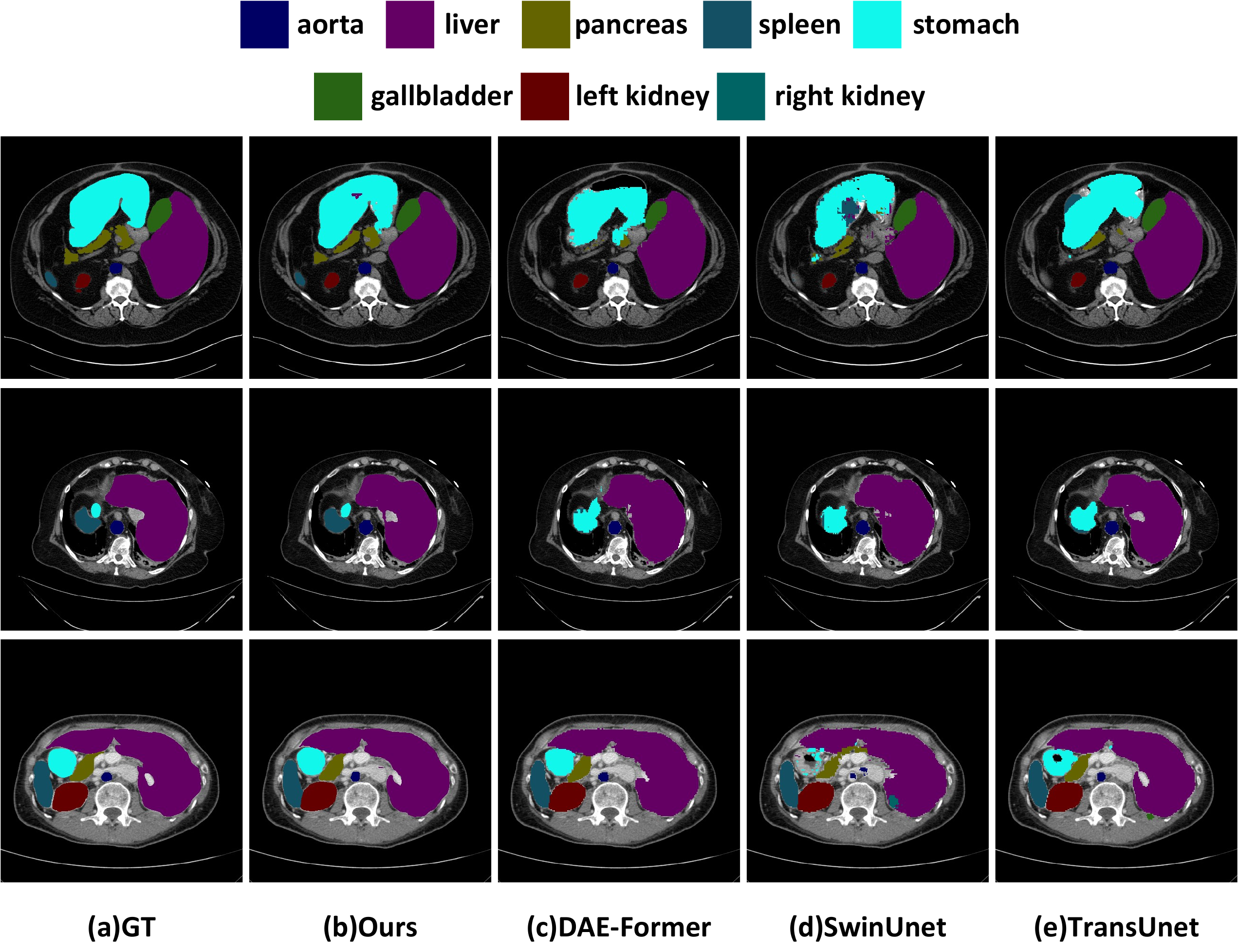}
\end{center}
   \caption{The qualitative comparisons between SAMed and the SOTA methods, including TransUnet~\cite{chen2021transunet}, SwinUnet~\cite{swinunet} and DAE-Former~\cite{azad2023daeformer}.}
\label{fig:qc}
\end{figure*}

\subsubsection{Qualitative comparison.} We compare the segmentation masks between SAMed and current state-of-the-art methods including TransUnet~\cite{chen2021transunet}, SwinUnet~\cite{swinunet} and DAE-Former~\cite{azad2023daeformer} qualitatively. The comparison result is shown in Fig.~\ref{fig:qc}. Compared with other methods, the segmentation regions predicted by SAMed are more smooth and correct. We attribute this phenomenon to the powerful feature extraction ability of the large-scale SAM model and the reasonable finetuning strategies adopted by SAMed, which benefit the customization of SAM.

\subsection{Ablation study}
\subsubsection{Lora finetuning in SAMed.} Compared with the traditional finetuning on the segmentation head, SAMed finetunes not only the mask decoder but also the image encoder with LoRA. Tab.~\ref{fm} shows such strategy can lead to higher segmentation accuracy. We think the data domain of SAM is mainly nature image, which differs a lot from the medical images. Therefore, the customization of the image encoder helps SAMed to extract meaningful features from the medical images, which benefits the subsequent processing in the mask decoder.

\begin{table*}[t!]
\centering
\caption{Ablation study on finetuning method of SAMed (The embedding in prompt encoder is finetuned by default).}
\footnotesize
\resizebox{\textwidth}{!}{
\begin{tabular}{c|c|cccccccc}
\toprule
Methods & DSC$\uparrow$ & Aorta& Gallbladder& Kidney(L)& Kidney(R)& Liver& Pancreas& Spleen& Stomach\\
\midrule
Mask decoder & 67.95 & 79.94 & 39.49 & 76.72 & 73.55 & 90.87 & 44.15 & 73.79 & 65.11 \\
Image encoder + mask decoder & \textbf{81.88} & \textbf{87.77} & \textbf{69.11} & \textbf{80.45} & \textbf{79.95} & \textbf{94.80} & \textbf{72.17} & \textbf{88.72} & \textbf{82.06} \\
\bottomrule
\end{tabular}
}
\label{fm}
\end{table*}

The mask decoder of SAM also contains a light-weight transformer layer to decode the extracted image tokens. We also explore the performance variance when we finetune this transformer layer together with the image encoder with LoRA. Since the resulting updated parameter shape is obviously smaller than the original SAMed, we denote this version as SAMed\_s. The performance comparison between SAMed and SAMed\_s is shown in Tab.~\ref{LoRA_decoder}. The performance of SAMed\_s is not as good as SAMed, which means all the parameters in the mask decoder should be updated during customization process. However, since SAMed\_s contains fewer updated parameters, the overhead in deployment and storage of SAMed\_s is smaller. If the practical usage has a strict limit over the deployment and storage, SAMed\_s can become a substitute of SAMed.

\begin{table*}[t!]
\centering
\caption{Ablation study on applying LoRA finetuning to the transformer of mask decoder.}
\footnotesize
\resizebox{\textwidth}{!}{
\begin{tabular}{c|c|c|cccccccc}
\toprule
Methods & DSC$\uparrow$ & Model size & Aorta& Gallbladder& Kidney(L)& Kidney(R)& Liver& Pancreas& Spleen& Stomach\\
\midrule
SAMed & \textbf{81.88} & 18.81M & \textbf{87.77} & \textbf{69.11} & \textbf{80.45} & \textbf{79.95} & \textbf{94.80} & \textbf{72.17} & \textbf{88.72} & \textbf{82.06} \\
SAMed\_s & 77.78 & \textbf{6.32M} & 83.62 & 57.11 & 79.63 & 78.92 & 93.98 & 65.66 & 85.81 & 77.49 \\
\bottomrule
\end{tabular}
}
\label{LoRA_decoder}
\end{table*}

Tab.~\ref{rank_size} presents the performance of SAMed when we adjust the rank size in the LoRA layers. We discover the performance of SAMed increases steadily in a certain rank range, but the performance drops significantly when the rank is too large. We think the LoRA layers in SAMed need the basic amount of parameters to accomplish the customization process on medical image dataset, but too many trainable parameters may undermine the ability of SAMed to adopt the original capabilities of SAM in image segmentation, which increases the training difficulty.

\begin{figure*}[t]
\begin{center}
\includegraphics[width=\linewidth]{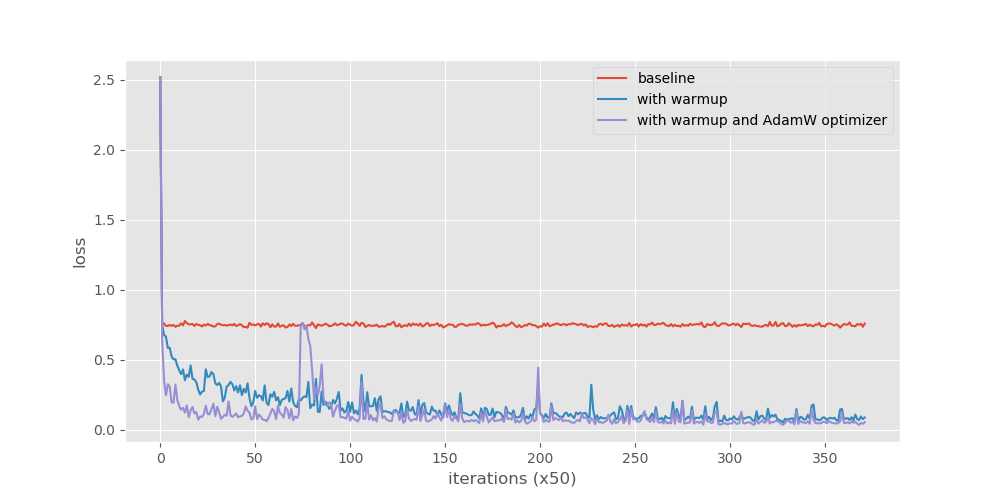}
\end{center}
   \caption{The comparisons of loss curves between different training strategies. ``baseline" indicates the SAMed model trained without warmup and AdamW optimizer (we use SGD optimizer instead). The result illustrates warmup can lead to successful convergence of SAMed, and the AdamW optimizer can further reduce the training loss in the later stage.}
\label{fig:loss_curves}
\end{figure*}

\begin{table*}[t!]
\centering
\caption{Ablation study on the rank size on the LoRA layer}
\footnotesize
\resizebox{\textwidth}{!}{
\begin{tabular}{c|c|cccccccc}
\toprule
Rank size & DSC$\uparrow$ & Aorta& Gallbladder& Kidney(L)& Kidney(R)& Liver& Pancreas& Spleen& Stomach\\
\midrule
1 & 78.26 & 81.86 & 64.54 & \textbf{81.97} & 81.18 & 93.79 & 60.80 & 88.33 & 73.64 \\
4 & \textbf{81.88} & \textbf{87.77} & \textbf{69.11} & 80.45 & \textbf{79.95} & \textbf{94.80} & \textbf{72.17} & \textbf{88.72} & \textbf{82.06} \\
16 & 69.03 & 78.74 & 55.54 & 71.98 & 65.08 & 91.38 & 45.01 & 81.39 & 63.11 \\
\bottomrule
\end{tabular}
}
\label{rank_size}
\end{table*}

Finally, we study the recipe of integrating LoRA to the projection layers in the original transformer blocks in SAMed and report the results in Tab.~\ref{proj}. Our recipe to adopt LoRA to q and v projection layers is reasonable and enjoys the best performance among the candidates. If we adopt LoRA to all of the projection layers (q,k,v and o), the performance drops significantly, which confirms our argument: too much LoRA customization may undermines the ability of SAMed to utilize the intrinsic image segmentation power of SAM, which brings extra obstacles during customization process.

\begin{table*}[t!]
\centering
\caption{Ablation study on the LoRA applied to different projection layers}
\footnotesize
\resizebox{\textwidth}{!}{
\begin{tabular}{c|c|cccccccc}
\toprule
Proj layer & DSC$\uparrow$ & Aorta& Gallbladder& Kidney(L)& Kidney(R)& Liver& Pancreas& Spleen& Stomach\\
\midrule
Q & 73.76 & 84.73 & 58.68 & 76.25 & 70.58 & 91.16 & 53.69 & 83.54 & 71.47 \\
Q+V & \textbf{81.88} & \textbf{87.77} & \textbf{69.11} & \textbf{80.45} & \textbf{79.95} & \textbf{94.80} & \textbf{72.17} & \textbf{88.72} & \textbf{82.06} \\
Q+K+V+O & 50.93 & 46.51 & 46.92 & 51.38 & 47.33 & 86.55 & 22.23 & 64.98 & 41.51 \\
\bottomrule
\end{tabular}
}
\label{proj}
\end{table*}

\subsubsection{The effect of training strategies.} In this part, we mainly verify the efficacy of warmup and AdamW optimizer in the customization process of SAMed. As shown in Tab.~\ref{ts}, these training strategies brings significant performance boost for SAMed. We show the loss curves in Fig.~\ref{fig:loss_curves}. With the training strategies, the training process of SAMed stabilizes a lot and can converge to a obviously lower loss value. 

\begin{table*}[t!]
\centering
\caption{Ablation study on the training strategies of SAMed.}
\footnotesize
\resizebox{\textwidth}{!}{
\begin{tabular}{c|c|cccccccc}
\toprule
Training strategies & DSC$\uparrow$ & Aorta& Gallbladder& Kidney(L)& Kidney(R)& Liver& Pancreas& Spleen& Stomach\\
\midrule
No strategies & 56.54 & 57.28 & 31.60 & 67.58 & 58.14 & 86.66 & 26.02 & 75.27 & 49.79 \\
+warmup & 75.29 & 81.45 & 61.23 & \textbf{81.67} & 77.68 & 92.51 & 53.29 & 85.92 & 68.57 \\
+warmup+AdamW & \textbf{81.88} & \textbf{87.77} & \textbf{69.11} & 80.45 & \textbf{79.95} & \textbf{94.80} & \textbf{72.17} & \textbf{88.72} & \textbf{82.06} \\
\bottomrule
\end{tabular}
}
\label{ts}
\end{table*}

What's more, we find the training strategies can even benefit other medical image segmentation methods. We adopt two representative methods - TransUnet~\cite{chen2021transunet} and SwinUnet~\cite{swinunet} to demonstrate this argument. We also freeze all the transformer layers in TransUnet and SwinUnet with LoRA and customize these models on Synapse dataset (We can do this because TransUnet and SwinUnet are adapted from the corresponding pretrained models~\cite{dosovitskiy2021an,liu2021Swin} on ImageNet~\cite{deng2009imagenet}). Tab.~\ref{ts_analysis} provides the performance of the original models, LoRA finetuned models and LoRA finetuned models with our training strategies. Compared with the original models, the performance of LoRA finetuned models drops a lot. This phenomenon is reasonable because the pretrained models used in these methods are not trained for image segmentation specifically. With our training strategies, the performance of these methods increases significantly (except HD of TransUnet), which demonstrates the validity of our training strategies. These strategies may potentially benefit to other models with LoRA finetuning, and we will explore this theme in the future work.


\begin{table*}[t]
\begin{center}
\caption{Ablation study about our training strategies on other medical image segmentation methods.}
\label{ts_analysis}
\setlength{\tabcolsep}{1mm}{
\begin{tabular}{ccccc}
\toprule
\multirow{2}{*}{Method} & \multicolumn{2}{c}{TransUnet~\cite{chen2021transunet}} & \multicolumn{2}{c}{SwinUnet~\cite{swinunet}} \\ \cmidrule(r){2-3} \cmidrule(r){4-5}
& DSC$\uparrow$ & HD$\downarrow$ & DSC$\uparrow$ & HD$\downarrow$ \\ \midrule
Original & 77.78 & 31.69 & 79.13 & 21.55 \\
+LoRA finetune & 74.67 & 33.52 & 69.61 & 41.47 \\
+Our strategies & 77.49 & 44.94 & 71.22 & 31.24 \\
\bottomrule
\end{tabular}}
\end{center}
\end{table*}

\section{Conclusion}
In this paper, we explore the feasibility of customizing the large-scale model to medical image segmentation and demonstrate the customized large-scale model with suitable customization strategies can achieve highly competitive and remarkable results compared with the previous well-designed medical image segmentation models. We select SAM as the large-scale model due to its outstanding segmentation capability on openset images and the multi-organ segmentation as the target task for its ubiquitous application in computer-assisted medical diagnosis. Our method, SAMed, which equips with LoRA adapted image encoder and the suitable training strategies, is capable of achieving competitive results on Synapse dataset. Moreover, SAMed is fully compatible with SAM, and the deployment and storage overhead of SAM on the segment anything system is quite marginal, which is acceptable for practical usage.


%
%
%
\bibliographystyle{splncs04}
\bibliography{egbib}
%




\end{document}